\documentclass[acmsmall,screen, review=false, timestamp=false, nonacm, authorversion]{acmart}

\usepackage{subcaption}
\usepackage{wrapfig}
\usepackage{booktabs}
\usepackage{placeins}

\AtBeginDocument{%
  }

\setcopyright{acmcopyright}
\copyrightyear{2018}
\acmYear{2018}
\acmDOI{XXXXXXX.XXXXXXX}

\begin{document}

\title{ChatGPT Prompting Cannot Estimate Predictive Uncertainty in High-Resource Languages}

\author{Martino Pelucchi}
\email{m.pelucchi@student.rug.nl}
\author{Matias Valdenegro-Toro}
\orcid{0000-0001-5793-9498}
\email{m.a.valdenegro.toro@rug.nl}
\affiliation{%
  \institution{Department of AI, University of Groningen}
  \streetaddress{Nijenborgh 9}
  \city{Groningen}
  \country{The Netherlands}
  \postcode{9747AG}
}

\renewcommand{\shortauthors}{Pelucchi and Valdenegro.}

\begin{abstract}
    ChatGPT took the world by storm for its impressive abilities. Due to its release without documentation, scientists immediately attempted to identify its limits, mainly through its performance in natural language processing (NLP) tasks. This paper aims to join the growing literature regarding ChatGPT’s abilities by focusing on its performance in high-resource languages and on its capacity to predict its answers’ accuracy by giving a confidence level. The analysis of high-resource languages is of interest as studies have shown that low-resource languages perform worse than English in NLP tasks, but no study so far has analysed whether high-resource languages perform as well as English. The analysis of ChatGPT’s confidence calibration has not been carried out before either and is critical to learn about ChatGPT's trustworthiness. In order to study these two aspects, five high-resource languages and two NLP tasks were chosen. ChatGPT was asked to perform both tasks in the five languages and to give a numerical confidence value for each answer. The results show that all the selected high-resource languages perform similarly and that ChatGPT does not have a good confidence calibration, often being overconfident and never giving low confidence values.
\end{abstract}

\begin{CCSXML}
<ccs2012>
<concept>
<concept_id>10010147.10010341.10010349.10010345</concept_id>
<concept_desc>Computing methodologies~Uncertainty quantification</concept_desc>
<concept_significance>500</concept_significance>
</concept>
<concept>
<concept_id>10010147.10010178.10010216</concept_id>
<concept_desc>Computing methodologies~Philosophical/theoretical foundations of artificial intelligence</concept_desc>
<concept_significance>500</concept_significance>
</concept>
<concept>
<concept_id>10002944.10011123.10010912</concept_id>
<concept_desc>General and reference~Empirical studies</concept_desc>
<concept_significance>500</concept_significance>
</concept>
</ccs2012>
\end{CCSXML}

\ccsdesc[500]{Computing methodologies~Uncertainty quantification}
\ccsdesc[500]{Computing methodologies~Philosophical/theoretical foundations of artificial intelligence}
\ccsdesc[500]{General and reference~Empirical studies}

\keywords{Large Language Models, Uncertainty Estimation, ChatGPT.}

\received{20 February 2007}
\received[revised]{12 March 2009}
\received[accepted]{5 June 2009}

\maketitle

\section{Introduction}
\label{sec:introduction}

ChatGPT is a large language model developed by OpenAI \footnote{ \url{https://openai.com/}} based on their previous model GPT-3/3.5 As a generative pre-trained model, GPT-3 was created by training an extremely large neural network (175 billion parameters) through deep learning on enormous amounts of data \citep{brown2020language}. Thanks to its generative nature it is able to create coherent text starting from an initial input. ChatGPT is a fine-tuned version of this model specialising in following instructions and holding conversations. Moreover, it was also trained using reinforcement learning from human feedback \citep{christiano2017deep}, which leads to more accurate and faster learning thanks to improved reward functions. 

Ever since its release on the 30th of November 2022, ChatGPT has become extremely popular due to its accessibility and ability to provide answers to all sorts of requests. In the first week, the chatbot developed by OpenAI reached a million users and as of January 2023, it surpassed 100 million users \citep{ruby-2023}. It is until now freely available and has a very straightforward user interface. As it gained popularity, users quickly realised its impressive natural language processing (NLP) and multilingual abilities, but as the model was released without documentation, and its limits were not clear. To better understand ChatGPT's strengths and weaknesses, researchers started to analyse and study the model's performance from the scientific point of view.

Although ChatGPT's language processing abilities have been studied through NLP tasks, very little research has been carried out on whether non-English high-resource languages show the same level of performance as English \cite{adewumi2022itakuroso}. The research on ChatGPT's multilingual abilities has focused on Machine Translation, which does not lead to any conclusions on its abilities to perform NLP tasks in different languages.  Considering ChatGPT was released with little to no documentation, research is the only approach to discovering the limits and strengths of its linguistic abilities. It is therefore necessary to perform thorough testing and analyse ChatGPT's abilities in different languages. 

LLMs like ChatGPT often produce "hallucinations" \cite{zhang2023siren}, incorrect and nonsensical predictions, and there is public interest in detecting them \cite{bender2021dangers}. Uncertainty estimation is one possible way to detect and signal to the user that predictions are unreliable, which closely relates to out of distribution detection \cite{yang2021generalized}.

\begin{figure}[!t] 
    \centering
    \begin{subfigure}{0.75\textwidth}
        \includegraphics[width=\linewidth]{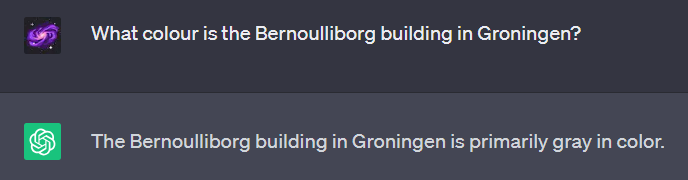}
        \caption{ChatGPT Prompt and Answer}
    \end{subfigure}
    \begin{subfigure}{0.24\textwidth}
        \includegraphics[width=\linewidth]{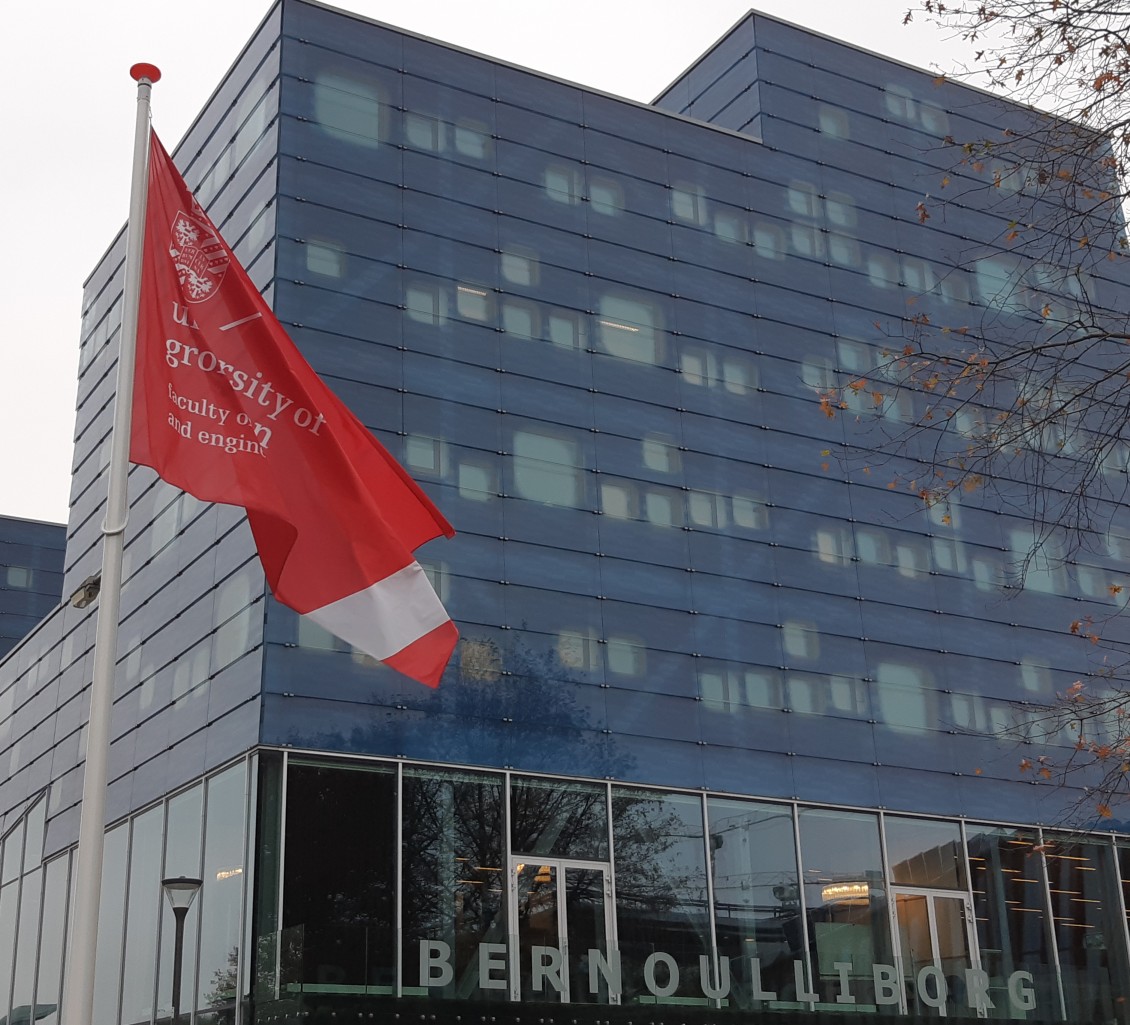}
        \caption{Bernoulliborg}
    \end{subfigure}
    
    \caption{ChatGPT answering with an incorrect fact rather than admitting its lack of knowledge on the topic. The building in question is primarily blue as shown on the right.}
    \label{ignorant}
\end{figure}

Another aspect of ChatGPT that has not yet been studied is its confidence, namely whether it is able to give a reliable confidence value for its answers, one that matches their accuracy. The study of ChatGPT's confidence calibration is important due to the blind trust many users have in its answers. Simple interactions with the chatbot show that it often gives inaccurate answers rather than acknowledging its ignorance on the topic, as shown in Figure \ref{ignorant}.

Accurate confidence estimation is necessary for trustworthiness \cite{hasan2023survey} and as direct feedback to the user \cite{grote2021trustworthy}, to answer as simple question: is the model guessing or is it confident on its answer? This is required for many applications and uses of chatbots. Without proper confidences, users might rely on a guess instead of discarding an answer and consulting another (better) resource.

When properly prompted, ChatGPT is able to give a level of confidence in its answer, but no study has analysed whether its perception is accurate. Discovering whether ChatGPT gives a reliable confidence value on its answers would be a step forward towards being able to recognise which answers can be trusted.

This paper aims to fill the literature gap by comparing ChatGPT's accuracy and confidence across multiple high-resource languages in two NLP tasks. In doing so, it strives to answer two research questions: Is ChatGPT equally  accurate in high-resource languages? Is ChatGPT's confidence in its answers well-calibrated?

The contributions of this work are: we define and test prompts to extract confidence estimates directly in ChatGPT's answer for sentiment analysis and common sense reasoning. We evaluate ChatGPT's prompting confidence estimation capabilities through calibration in sentiment analysis and common sense reasoning, and find that ChatGPT produces very uncalibrated confidence estimates in its answers through a prompt.

In the following sections the tasks, languages and data will be discussed, as well as the setup of the experiment. Afterwards, the results will be presented and discussed. Lastly, conclusions will be drawn.

\section{Related Work}

ChatGPT has been compared to a multitude of models, from fine-tuned to general ones, in a variety of NLP tasks and languages. \cite{qin2023chatgpt} used 7 different tasks (reasoning, natural language inference, question answering, dialogue, summarization, named entity recognition, and sentiment analysis) to compare ChatGPT to GPT 3.5 and to models fine-tuned to the tasks. They found that ChatGPT performed slightly worse than fine-tuned models in the majority of tasks, but its performance was better than GPT 3.5 for the majority of tasks except for reasoning tasks like common sense questions. 

\cite{jiao2023chatgpt} analysed its multilingual abilities through translation tasks between English, German, Chinese, and Romanian. The results indicated that ChatGPT performs similarly to popular machine translation models like Google Translate for high-resource languages such as English and German, but falls behind when translating into Romanian, a lower-resource language.

In order to study ChatGPT's NLP abilities and language knowledge \cite{bang2023multitask} performed a multilingual and multitask evaluation. They performed 7 NLP tasks (summarisation, machine translation, sentiment analysis, question answering, misinformation detection, task-oriented dialogue, open domain kgd) across a set of languages with mixed resource levels (English, Chinese, French, Indonesian, Korean, Javanese, Sundanese, Buginese). Although the results slightly varied depending on the task, the general finding was that ChatGPT achieved strong performances in high- and medium-resource languages (e.g. English, Indonesian) while it struggled to understand low-resource languages (e.g. Javanese, Sundanese).

These papers indicate that although ChatGPT can understand a vast number of languages, its performance on NLP tasks is dependent on the amount of resources it has at its disposal for a specific language. In order to study ChatGPT's performance when executing tasks in different languages, it is therefore necessary to look at their resource levels. To do so, the spread of languages in ChatGPT's training data must be investigated.

Evaluation of ChatGPT in terms of confidence and uncertainty is sparse. \cite{li2023evaluating} evaluates ChatGPT's uncertainty through prompts in information extraction tasks, while \cite{kumar2023conformal} builds LLMs with predictive uncertainty by applying conformal prediction.

\subsection{Linguistic Resources}

\begin{wraptable}{r}{0.5\textwidth}
    \caption{Distribution of the most popular European languages in web content \citep{ani-2023} and in the Common Crawl dataset \citep{commoncrawl-no-date} in percentage.}
    \label{tab:languages}
    \begin{tabular}{lll} \hline
        \centering 
        Languages & Web Content & Common Crawl \\\hline
        English & 58.8\% &  46.3\% \\ 
        Spanish & 4.3\% & 4.6\% \\ 
        German & 3.7\% & 5.9\% \\ 
        French & 3.7\% & 4.7\% \\ 
        Italian & 1.6\% & 2.7\% \\ 
    \end{tabular}

\end{wraptable}

OpenAI keeps the data that was used to train any of their GPT models confidential; it is therefore impossible to know exactly the percentage of texts in each language in the training data. Nonetheless, there are a few heuristics that can be used to get an educated guess on the matter. Although the specifics are not public, it is generally thought that OpenAI gets its datasets from crawling the internet, but not much information is known. Consequently, statistics such as the distribution of languages in web content \citep{ani-2023} or in the Common Crawl dataset \citep{commoncrawl-no-date} can give an estimation of the distribution of language in ChatGPT's training data. 

Table \ref{tab:languages} gives an overview of the percentage of web content and of the Common Crawl dataset in different European languages. Although, for example, 1.6\% might seem like a very small percentage, all the languages listed are still commonly considered high-resource, which is usually defined as languages with over 1\% of web content. In fact, ChatGPT seems fluent in all five languages. Nevertheless, the disparity between English and other languages is still notable. This begs the question of whether a more in-depth analysis would show that the disparity in language distribution leads to a significant difference in performance.

\section{Estimating ChatGPT's Confidence Estimation Capabilities}
\label{sec:methods}

\subsection{Tasks and Languages}
In order to study ChatGPT's accuracy and confidence in high-resource languages, it was decided to compare its performance in two NLP tasks and five languages. The tasks were sentiment analysis (SA) and common sense reasoning (CSR), while the languages were English, Spanish, German, French and Italian.

Sentiment analysis is a task whose goal is to find opinions and identify the sentiment they carry, whether positive, negative, or neutral \citep{medhat-2014}. Common sense reasoning is the task of making inferences or answering questions based on common sense knowledge. This knowledge is considered to be the set of facts that reflect the general understanding of the world and of humans \citep{davis-2015}. 
These tasks were chosen as they involve two skills that are fundamental for conversation and consequently for chatbots like ChatGPT.

Previous literature was analysed to decide which languages to use in this study. Since the only previous research which performed a multilingual and multitask analysis of ChatGPT focused on Asian languages \citep{bang2023multitask}, European languages were chosen as the focus of this study. The decision to use the most popular languages was based on multiple factors. The first one is the number of speakers. Using the most popular languages means that the findings of this paper can affect a high number of people and their usage of ChatGPT. The second factor is that there is more uncertainty around non-English high-resource languages. While the drop in performance for low-resource languages is obvious, it is not clear whether ChatGPT's performance in other high-resource languages is as reliable as in English. In fact, although all five languages are considered high-resource, there is still a notable difference in resources between the languages.

\textbf{Datasets}. To study ChatGPT's performance in different languages it was necessary to find multilingual datasets for the two tasks of sentiment analysis and common sense reasoning. To this end, Hugging Face was used. Hugging Face\footnote{\url{https://huggingface.co/}} is an AI Community that promotes open-source contributions and acts as a hub for models and datasets. The datasets can be filtered based on tasks and languages and this option was used to select the two datasets for this project.

In order to study how ChatGPT performs in SA, the Unified Multilingual Sentiment Analysis Benchmark (UMSAB) \citep{barbieri-etal-2022-xlm} was chosen. This multilingual dataset was constructed by combining 8 monolingual sentiment analysis datasets. All datasets are made of 3033 tweets evenly split between the three possible labels: "positive", "negative" and "neutral".

For CSR, the XCSR dataset \citep{lin2021common} was chosen. This dataset was created to improve multi-lingual language models in CSR. It consists of common sense multiple-choice questions with 5 possible answers in 16 languages. The dataset is split into two sub-datasets: X-CODAH and X-CSQA.
The validation split of X-CSQA was chosen, as it is the only part which contains all the necessary fields. For each language, it contains 1000 datapoints.

\textbf{Preprocessing.} Although the two original datasets are already curated, some preprocessing steps were needed for the scope of this project. 

Firstly, as this experiment did not envisage a validation phase, the data splits were merged together, in order to have a larger pool of sentences and questions.
Secondly, in the UMSAB dataset, some sentences contained Unicode characters that had not been converted. All the sentences containing a Unicode character were excluded to remove any possible confusion for ChatGPT. 

Thirdly, some sentences and questions around topics such as sex and drugs caused ChatGPT to give warnings regarding its content policy. It was not possible to know which sentences would violate the content policy and to remove them beforehand automatically. Therefore, those sentences or questions were removed manually as they were found.

Lastly, as the datapoints needed to be manually fed to ChatGPT in small batches, the size of the datasets was reduced. For every language and task, a subset of 300 datapoints was randomly sampled. For the sentiment analysis task, the subset was kept balanced.

\subsection{Prompt Engineering} \label{sec:_prompts}

As a chatbot, the only way to interact with ChatGPT is through prompts. Therefore, a prompt needed to be established for each task. The prompt had to be defined carefully since it had multiple roles. Firstly, the task needed to be explained. For this experiment, this meant specifying how to classify the sentiment of tweets or how to answer multiple-choice questions. Secondly, the format of the input needed to be explained. This was done to ensure that ChatGPT knew what to expect and what to do with the tweets or questions. Lastly, the output desired from ChatGPT needs to be defined. This was done to allow for an automatic scanning and analysis of the responses.

The process of creating the prompt started with simple requests to perform sentiment analysis or answer multiple-choice questions. ChatGPT immediately showed an understanding of the tasks, but the format of its response varied. The answer was often explained and the confidence was not expressed numerically. This led to the addition of a sentence detailing the desired output. After noticing that an in-depth explanation of the task did not seem to influence the answers, the focus shifted to the number of datapoints that could be sent together (a block).

The goal was to maximise the datapoints per block to speed up the experiment process. At around 20 tweets per block ChatGPT would often ``forget" the task and simply describe the tweets (e.g. saying that they are a collection of tweets). To avoid this issues, a sentence describing the block that was to come was added and its size was restricted to 15 datapoints. 

Lastly, the English prompt was translated by a professional translator into the different languages in an attempt to improve the performance in non-English languages. A small experiment was conducted using sentiment analysis to compare the performance of ChatGPT using the English versus the translated prompts. Since a significant difference in accuracy was found for some languages (see Appendix \ref{mcnemar}), the translated prompt was used for all languages and tasks.

After testing out different variations, the final English prompts that were used for the experiment were the following:
\begin{itemize}
    \item \textbf{Sentiment Analysis:} Hello, I would like you to perform sentiment analysis on 15 English sentences. Please classify each sentence as “Positive”, “Negative”, or “Neutral” according to the sentiment it expresses. Moreover, please rate your confidence in the answer you gave between 0 and 100\%. The answer should be a list with the format “Sentiment (confidence\%)”. Say “understood” if you have understood. The list of sentences will follow.
    \item \textbf{Common Sense Reasoning:} Hello, I would like you to answer 15 common sense multiple-choice questions in English. For each question please give the letter that corresponds to the correct answer. Moreover, please give your confidence in the answer you gave between 0 and 100\%. The answer should be a list with the format “Letter (confidence\%)”. Say “understood” if you understand. The list of questions will follow.
\end{itemize}
A complete list of all the prompts used in the different languages for both tasks can be found in Appendix \ref{translated}. We also produce and evaluate Zero Shot Chain of Thought \cite{kojima2022large} Prompts by adding the sentence \textit{"Please
explain your reasoning step by step"}. These prompt details are available in Appendix \ref{CoTPrompt}.

\subsection{Exploratory Analysis}\label{sec:_exploratory}
Once the prompt was defined, an exploratory analysis was conducted.
Its goal was to identify patterns in the responses as well as any confounding variables.

A first finding was that, after the prompt, the number of blocks that could be sent consecutively was not consistent. Sometimes ChatGPT would perform the task on 3 blocks, while in other cases it stopped after the first block. It was therefore decided to send the prompt before each block.

A second finding was discovered when sending the same block multiple times. It was found that ChatGPT's answers varied, not only in confidence but also in the sentiment or answer chosen. Due to this discovery, it was decided to repeat the experiment 3 more times with a subset of the data in order to analyse the fluctuation in accuracy and confidence.

Another factor that was taken into account and tested is that ChatGPT's memory is chat-specific. The tests found that the same datapoints would sometimes lead to different answers even in the same chat. Nonetheless, in order to avoid any possible influence of the memory, a new chat was created for each language and for the repetition step of the experiment.

\subsection{Experimental Setup}
The experiment was run on OpenAI's website on the May 24, 2023 version of ChatGPT. The use of the website rather than the API was initially due to the non-existence of the API, and it was later preferred as most people use the web version of ChatGPT.

The process of gathering the answers was a repetition of the same two steps for each language and task. The prompt was sent and then a block of 15 sentences or questions followed. As there were 300 datapoints, the process was repeated 20 times per task and per language, every time in a new chat. The list of answers for each block was then copied into a text file in order to assemble the answers for all the blocks. An example of the ``conversation" can be seen in Figure \ref{convo}, while more examples from all tasks and languages can be found in Appendix \ref{examp}.

Afterwards, for every language and task, a subset of 75 datapoints was created. These subsets were sent 3 times each. The sentences and questions were presented in the same way to ChatGPT as previously explained.

\begin{figure}[!tbp] 
    \centering
    \includegraphics[width=7.5cm]{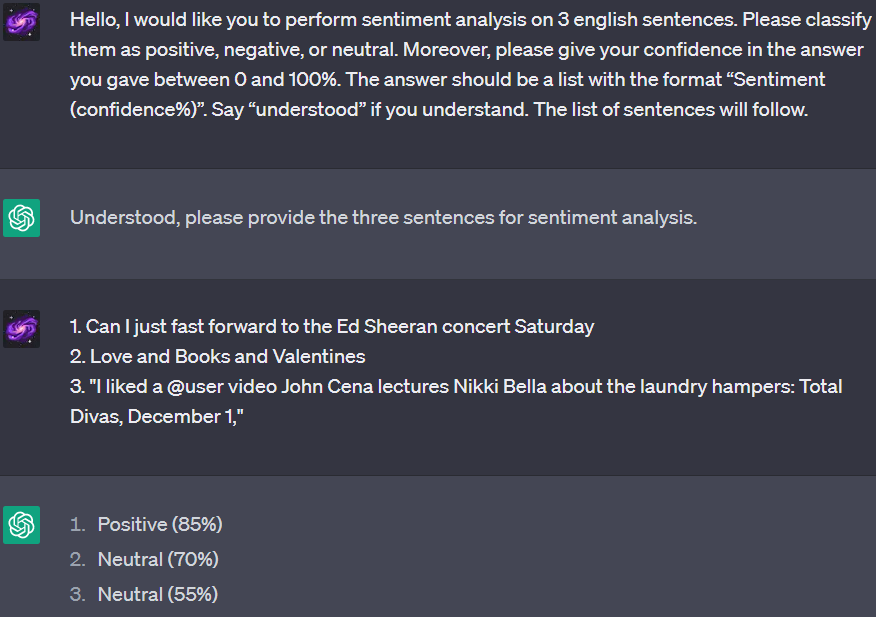}
    \caption{Showcase of the experiment procedure with ChatGPT. Number of datapoints lowered to 3 for demonstration purposes.}
    \label{convo}
\end{figure}

\subsection{Metrics}
To analyse the discrepancy between ChatGPT's confidence in its answers and their accuracy, the expected calibration error (ECE) and the maximum calibration error (MCE) were computed \citep{calibration}. The ECE is a weighted average of the absolute difference between the accuracy and the confidence of every bin and can be computed as follows:
\begin{equation}
    \text{{ECE}} = \sum_{m=1}^{M} \frac{|B_m|}{n} \bigg|\text{{acc}}(B_m) - \text{{conf}}(B_m)\bigg| \cdot 100,
\end{equation}
where $M$ is the number of bins, $B_m$ is the $m$th bin, $n$ is the number of datapoints and acc($B_m$) and conf($B_m$) are the average accuracy and confidence of the bin. As ECE uses a weighted average, the small range of confidence levels ChatGPT gives does not influence its value. If all bins were considered, the large number of empty bins would make the calibration error notably higher. The MCE is simply the biggest deviation between accuracy and confidence across all bins and can be computed as follows:
\begin{equation}
    \text{{MCE}} = \max_{m \in \{1,...,M\}} |\text{{acc}}(B_m) - \text{{conf}}(B_m)| \cdot 100.
\end{equation}
The two results for each language are displayed in Table \ref{tab:sent_anal}. The two metrics range from 0 to 100 and provide a standardised measure of calibration performance, with values closer to zero representing better calibration.

\section{Experimental Results}
\label{sec:results}

\subsection{Sentiment Analysis}

\begin{figure}[!t] 
    \centering
    \begin{subfigure}{0.32\textwidth}
        \includegraphics[width=\linewidth]{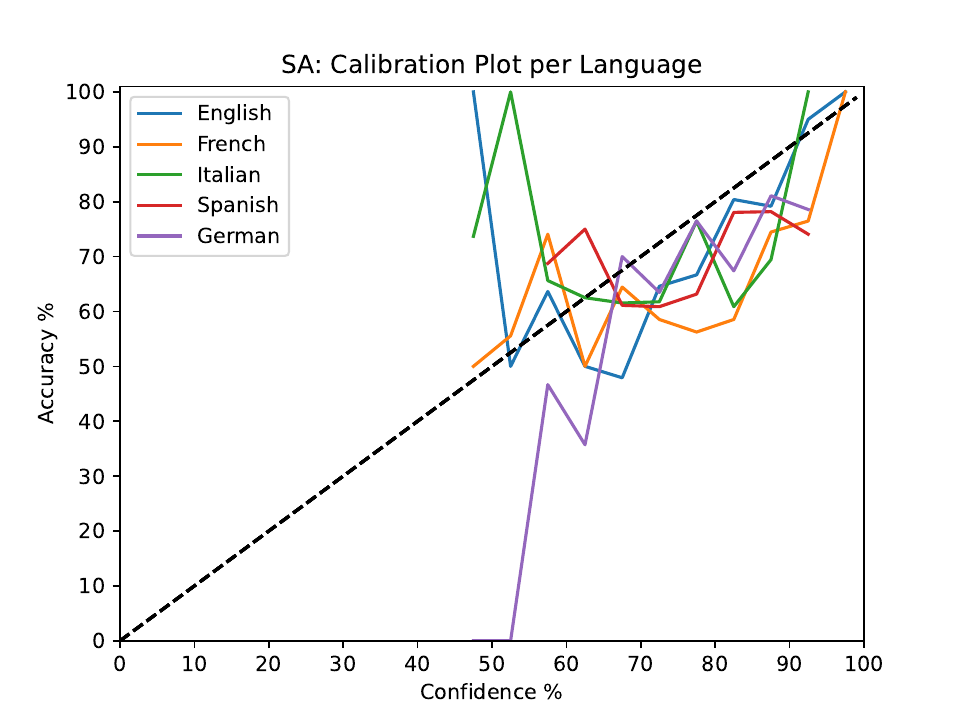}
        \caption{Calibration Plot}
        \label{sa_cal_plot}
    \end{subfigure}    
    \begin{subfigure}{0.32\textwidth}
        \includegraphics[width=\linewidth]{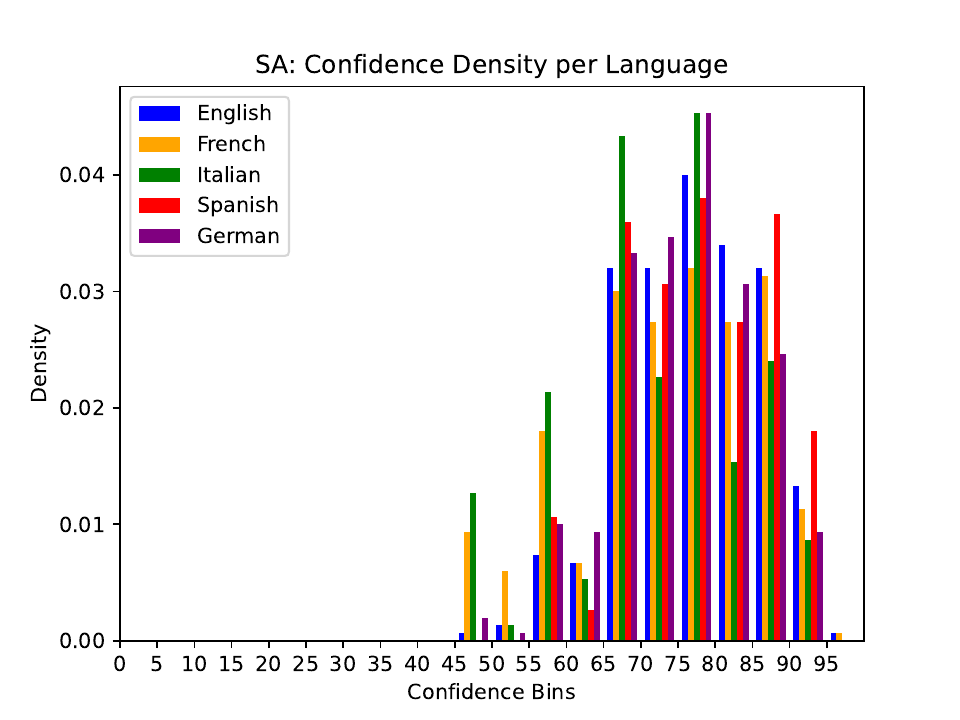}
        \caption{Confidence Histogram}
        \label{sa_conf_freq}
    \end{subfigure}    
    \begin{subfigure}{0.32\textwidth}
        \includegraphics[width=\linewidth]{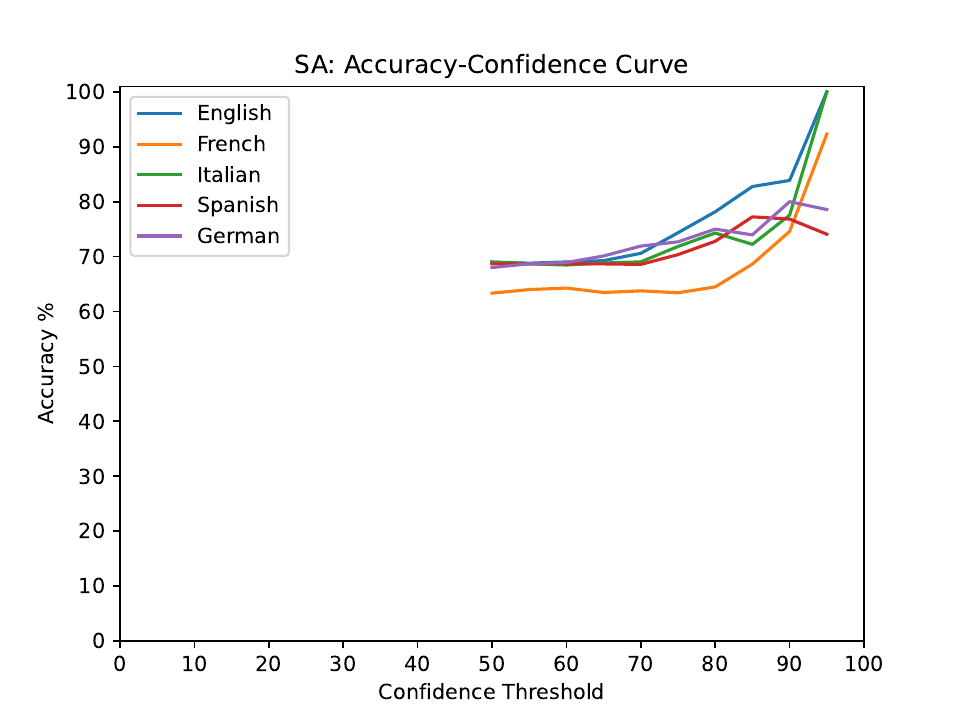}    
        \caption{Confidence-Accuracy Curve}
        \label{sa_conf_curve}
    \end{subfigure}        
    \caption{Visual results for sentiment analysis on all languages, comparing calibration, confidence distribution, and confidence-accuracy curves. ChatGPT is mostly overconfident in all languages, not giving a confidence level below 50\% in any language, increasing the confidence threshold has a minor effect on accuracy.}
    \label{sa_plots}
\end{figure}

\textbf{Qualitative Results}. In order to plot the results of the SA task, the data was grouped in 20 bins, each representing a 5\% confidence range. In Figure \ref{sa_cal_plot}, a calibration plot showcasing the accuracy per confidence bin of each language in the sentiment analysis task is presented. Before analysing it, the density of each confidence bin must be considered (see  Figure \ref{sa_conf_freq}). The first important result from the graphs is that ChatGPT does not give low confidence values, with the lowest being 50\% (60\% for Spanish) and the mode being 80\% across all languages. When focusing on the calibration plot, no language seems to perform better than the others, and it appears that all languages approximately follow the diagonal line representing a perfect calibration. Nonetheless, when declaring a confidence value between 50 and 60\%, ChatGPT appears underconfident (except for German), whereas between 60 and 90\% it appears overconfident. In the languages (English, French and Italian) that contain answers with confidence above 90\%, ChatGPT is appropriately confident.
Pairing these results with the density of the bins shows that the high accuracies for low confidences may be due to the low density in the concerned bins and that where most answers lie ChatGPT is overconfident.

Furthermore, an accuracy-confidence curve was generated to further analyse ChatGPT’s reliability, shown in Figure \ref{sa_conf_curve}. The curve was created by removing the data below an increasing threshold and computing the mean accuracy of the remaining data. It shows that, for all languages, the accuracy does not increase when the threshold increases from 50 to 70\%. In fact, the accuracy remains at around 65\% for French and 70\% for all other languages. This could be due to the general decrease in accuracy in that confidence range, as shown in Figure \ref{sa_cal_plot}. After the 70\% confidence threshold, all languages show some improvements as the threshold increases, with English and French having the biggest and smallest improvement respectively. Lastly, when the thresholds are set to 90 and 95\%, English, Italian and French show a steep increase in accuracy, while German and Spanish a slight decrease. 

\begin{table*}[t]
    \centering
    \caption{Summary of the results per language of ChatGPT in the sentiment analysis task. The values shown are mean accuracy and confidence (with standard deviation) as well as expected calibration error and maximum calibration error. There is no significant difference in accuracy between the languages.}
    \label{tab:sent_anal}
    \begin{tabular}{lllll}
        \toprule
        Language & Accuracy $\pm$ SD  & Confidence $\pm$ SD & ECE & MCE  \\
        \midrule
        English & 69.0 $\pm$ 1.5 & 79.3 $\pm$ 2.1  & 11.1 & 50.0\\
        French & 63.3 $\pm$ 2.4 & 76.4 $\pm$ 1.8  & 15.6 & 26.2\\
        Italian & 69.0 $\pm$ 2.0 & 74.9 $\pm$ 3.3  & 11.1 & 45.0\\
        Spanish & 68.7 $\pm$ 2.7 & 80.0 $\pm$ 1.6  & 12.6 & 21.0\\
        German & 68.0 $\pm$ 2.0 & 78.1 $\pm$ 1.8  & 10.1 &  55.0\\
        \bottomrule
    \end{tabular}
\end{table*}

\textbf{Quantitative Results}. The accuracy and confidence of ChatGPT in the SA task in the different languages can be found in Table \ref{tab:sent_anal}. English and Italian have the highest accuracy at 69\%, with Italian also having the lowest confidence. A chi-squared test of independence was performed in order to study the relationship between the language and the accuracy of ChatGPT. The test showed no significant association between the two variables, $\chi^2(4, N=1500)=3.208, p=0.52$.

To get a better understanding of the abilities of ChatGPT, its accuracy in the SA task was compared to that of two other models on the same dataset. The models, XLM-R \citep{conneau-etal-2020-unsupervised} and XLM-T \citep{barbieri-etal-2022-xlm}, are two large multilingual language models, with the second being fine-tuned on Twitter data. To have the best comparison, the F1 score of the two models when trained with all languages' training data was used. The F1 score was selected as it was the only metric presented for the two other models. It is important to remember that this is only a rough comparison as the ChatGPT results were only on a subset of the data. The F1 Score of the 3 models can be seen in Table \ref{tab:sa_compare}.

\begin{table}[!bp]
    \centering
    \caption{F1 score of ChatGPT in the sentiment analysis task compared to that of two benchmark models (XLM-R, XLM-T) on the UMSAB dataset. ChatGPT's performance is either equal or worse.}
    \label{tab:sa_compare}
    \begin{tabular}{llll}
        \toprule
        Language & ChatGPT   & XLM-R & XLM-T   \\
        \midrule
        English & 68.3 & 68.5 & 70.6\\ 
        French & 62.6 & 70.5  & 71.2\\ 
        Italian & 68.6 & 68.6 & 69.1 \\ 
        Spanish & 67.8 & 66.0  & 67.9\\ 
        German & 67.9 & 72.8 & 77.4\\ 
        \bottomrule
    \end{tabular}
\end{table}

From the results, it appears that change in performance partially depends on the language. All 3 models perform approximately the same in Italian and Spanish but ChatGPT performs worse than the other two in English, French and German. Overall, ChatGPT had a slightly worse performance compared to other benchmark multilingual language models when doing sentiment analysis. 

In order to assess the variance in ChatGPT's answers, the standard deviation of the accuracy and of the confidence across the 4 iterations of 75 datapoints were computed (Table \ref{tab:sent_anal}). The results show that there is slight variation both in the average accuracy and in the average confidence in all languages. Interestingly, English has the lowest accuracy deviation and the highest confidence deviation while for Spanish it is the opposite. It is important to consider that these results may vary if the questions are always asked in the same chat.

The ECE shows similar results across all languages, at around 13, with German having the lowest (10) and French having the highest (16). This means that on average ChatGPT's confidence levels are  10 to 16\% away from the accuracy. This confirms that ChatGPT has an overall decent calibration although not very accurate. The MCE varies more but has high results across all languages, ranging from 26 in French to 55 in German. This means that, although in general ChatGPT has a decent calibration, for some bins the values are completely wrong. This may also be due to the bins with little data having skewed results, as this metric is not influenced by the size of the bin.

\subsection{Common Sense Reasoning}

\begin{figure}[!t] 
    \centering
    \begin{subfigure}{0.32\textwidth}
        \includegraphics[width=\linewidth]{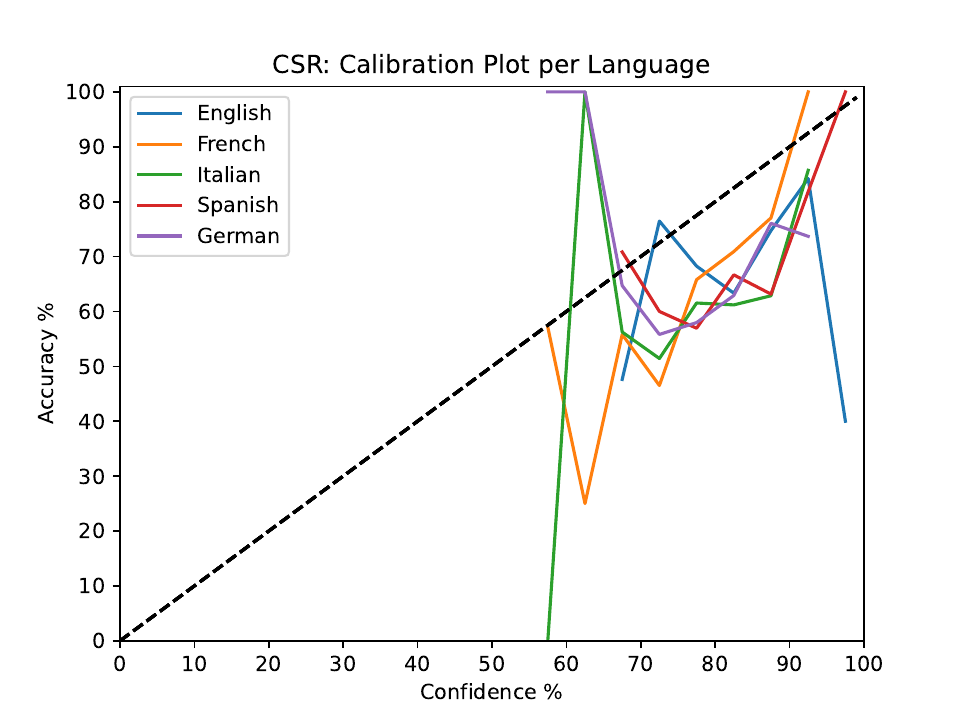}
        \caption{Calibration Plot}
        \label{csr_cal_plot}
    \end{subfigure}    
    \begin{subfigure}{0.32\textwidth}
        \includegraphics[width=\linewidth]{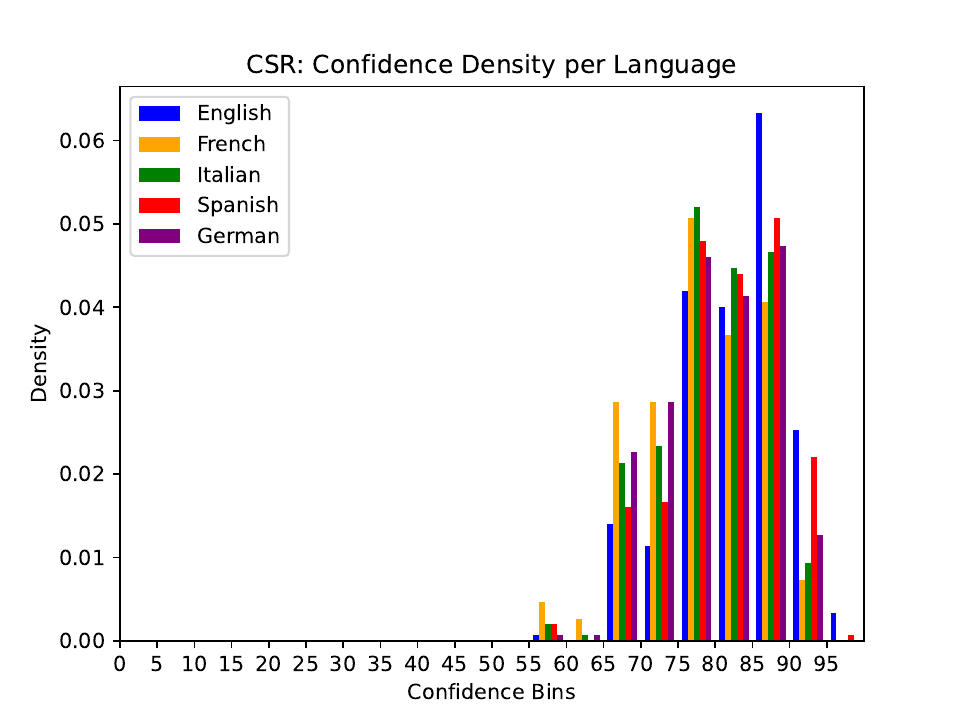}
        \caption{Confidence Histogram}
        \label{csr_conf_freq}
    \end{subfigure}    
    \begin{subfigure}{0.32\textwidth}
        \includegraphics[width=\linewidth]{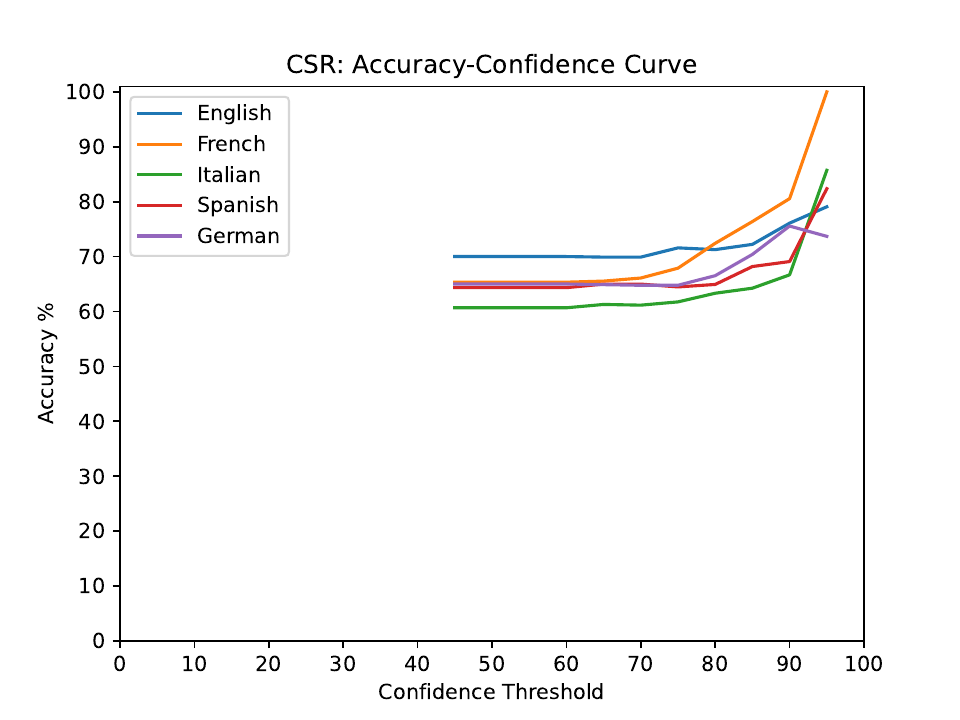}    
        \caption{Confidence-Accuracy Curve}
        \label{csr_conf_curve}
    \end{subfigure}
    \caption{Visual results for common sense reasoning on all languages, comparing calibration, confidence distribution, and confidence-accuracy curves. ChatGPT does not give any confidence levels below 55\% in any language, and is overconfident in all languages for the majority confidence bins. Increasing the confidence threshold has no strong effect on the accuracy.}
    \label{csr_plots}
\end{figure}

\textbf{Qualitative Results}. To analyse the results of ChatGPT in the CSR task, the same steps were taken as for the SA task. Figures \ref{csr_cal_plot} and \ref{csr_conf_freq} show similar trends to those of the first task. No confidences below 55\% are given and for the vast majority of the answers, between 65 and 90\% confidence, ChatGPT is overconfident, reaching accuracies between 50 and 70\%. Nonetheless, in that range, the data roughly follows the diagonal of perfect calibration. For confidences below 65\% and above 90\%, both extremely low accuracy and high accuracy values are seen, but this can be explained by the small size of the bins. In this task as well all languages seem to perform very similarly.

Further analysis performed with the accuracy-confidence curve found in Figure \ref{csr_conf_curve} shows that increasing the confidence threshold of the data included has no influence on the accuracy up until 75\%. In this interval, English has the highest accuracy at 70\% and Italian the lowest at 60\%. Once the threshold is increased past 75\%, the accuracy in French has a noticeable increase while the other languages only improve slightly until 90\%, at which point Spanish and Italian have a steep increase, German has a slight decrease and English continues on the same trend.

\textbf{Quantitative Results}. The accuracy and confidence of ChatGPT in the different languages while performing the CSR task can be found in Table \ref{tab:csr}. English has the highest accuracy (and confidence) while Italian has the lowest. It is also interesting to see that, although the mean confidence is higher than for the SA task, the accuracies are lower. A chi-squared test of independence was again performed in order to study the relationship between the language and the accuracy. The test showed no significant relationship between the two variables, $\chi^2(4, N=1500)=5.85, p=0.21$.

To understand how good ChatGPT is at answering CSR questions, its accuracy in the task was compared to that of two other models on the same dataset. The first model, XLM-R \citep{conneau-etal-2020-unsupervised}, is a large multilingual language model often used as a benchmark, while the second, MCP(XLM-R), is a variant of the first model which uses multilingual contrastive pre-training (MCP) \citep{lin2021common}. The results of the 3 models are shown in Table \ref{tab:csr_compare}.

\begin{table*}[t]
    \centering
    \caption{Summary of the results per language of ChatGPT in the common sense reasoning task. The values shown are mean accuracy and confidence (with standard deviation) as well as expected calibration error and maximum calibration error. There is no significant difference in accuracy between the languages.}
    \label{tab:csr}
    \begin{tabular}{lllll}
        \toprule
        Language & Accuracy $\pm$ SD  & Confidence $\pm$ SD & ECE & MCE  \\
        \midrule
        English & 70.0 $\pm$ 1.5 & 85.4 $\pm$ 2.2  & 15.8 & 60.0\\
        French & 65.3 $\pm$ 1.5 &  80.7 $\pm$ 2.0  & 15.7 & 40.0\\
        Italian & 60.6 $\pm$ 1.5 & 82.3 $\pm$ 1.1  & 21.8 & 60.0\\
        Spanish & 64.3 $\pm$ 3.1 & 83.9 $\pm$ 1.2  & 19.7 & 60.0\\
        German & 65.0 $\pm$ 2.5 & 82.4 $\pm$ 0.8  & 17.9 &  40.0\\
        \bottomrule
    \end{tabular}
\end{table*}

\begin{table}[!bp]
    \centering
    \caption{Accuracy of ChatGPT in the common sense reasoning task compared to that of two benchmark models (XLM-R, MCP(XLM-R)) on the X-CSQA dataset. ChatGPT's performance is better in all languages except Italian.}
    \label{tab:csr_compare}
    \begin{tabular}{llll}
        \toprule
        Language & ChatGPT & XLM-R & XLM-R$_{MCP}$\\
        \midrule
        English & 70.0 & 66.7 & 69.5\\ 
        French & 65.3 & 60.3 & 60.0\\ 
        Italian & 60.1 & 58.2 & 60.3 \\ 
        Spanish & 64.3 & 59.5  & 61.4\\ 
        German & 65.0 & 56.1 & 59.3\\ 
        \bottomrule
    \end{tabular}
\end{table}

Differently from SA, ChatGPT performs overall better than the two benchmark models in the CSR task. It has a higher accuracy in every language with the exception of Italian, for which the XLM-R model with MCP performs slightly better. These results suggest that the CSR task is harder than the SA task, with both ChatGPT and XLM-R performing worse, but also that ChatGPT is better at CSR than SA, at least compared to XLM-R.

In order to assess how much ChatGPT's answers vary in this task, the standard deviation was computed as for the sentiment analysis task. The results (Table \ref{tab:csr}) show again that both values vary similarly in all languages. With regards to accuracy, English and Italian have the lowest deviation for accuracy while Spanish has the highest. On the other hand, German has the lowest deviation for confidence while English has the highest.

As a way to further analyse the calibration of ChatGPT, the ECE and MCE were computed for each language and are presented in Table \ref{tab:csr}. 

The results are overall worse than in the SA task, with the lowest ECE being 16 for English and French and the highest being 21 for Italian. The analysis of the accuracy with the ECE seems to indicate that for the CSR task, English and Italian can be identified as the best and worst languages.
The MCE results for this task are also overall higher than for SA, although once again heavily dependent on the small bins.

Lastly, it is important to note that, differently from the first task, in the CSR task English has both the best accuracy and the best ECE, suggesting a potential overall better performance.

\subsection{Zero Shot Chain of Thought Prompting}

\begin{table}[t]
    \centering
    \caption{Effect of the Zero Shot Chain of Thought Prompts on five incorrect responses on Sentiment Analysis and Common Sense Reasoning tasks. It does not have a strong effect on the answers or confidence.}
    \label{tab:CoT}
    \begin{tabular}{lllll}
        \toprule
        Task & Changed Confidences & Confidence Change & Changed Answers & Accuracy Change \\
        \midrule
        SA & $3/5$ & -5\% & $1/5$ & +20\% \\ 
        CSR & $5/5$ & +11\% & $1/5$ & 0\%\\ 
        \bottomrule
    \end{tabular}
\end{table}

In an attempt to improve ChatGPT's performance in the two tasks, a small zero shot chain of thought experiment was run.
For this experiment, the original prompts were modified in order to be used for one sentence or question at a time. The 4 new prompts can be found in Appendix \ref{CoTPrompt}. For the data, from each task 5 datapoints for which ChatGPT got a wrong answer in the main experiment were randomly selected. To run the experiment a different chat was used for prompts with and without zero shot chain of thought as to avoid any possible influence. A summary of the results can be seen in Table \ref{tab:CoT} while the complete overview can be found in Appendix \ref{CoTResults}.

The results show that overall the zero shot chain of thought does not improve ChatGPT's performance in sentiment analysis and common sense reasoning. In fact, only once did it change ChatGPT's answer to a correct one.

Regarding confidence, they change in the majority of the answers, but in different directions for the two tasks. It appears that reasoning about sentiment analysis makes ChatGPT less confident in its answers, while the opposite happens for common sense reasoning. It is hard to understand the reason behind this but it might be an extension of the fact that overall the average confidence in the common sense reasoning task is higher than in the sentiment analysis task. This small experiment shows that zero shot chain of thought prompting might not improve ChatGPTs confidence estimation.

\section{Discussion and Analysis}
\label{sec:discussion}

\subsection{Performance in High-Resource Languages}

With regard to ChatGPT’s accuracy in performing NLP tasks in high-resource languages, no significant difference was found between languages. Moreover, except for English having the highest accuracy in both tasks, the other languages’ performance did not match their respective resource levels. For example, despite having the least resources of the five, Italian has the best accuracy in the SA task together with English. This suggests that ChatGPT can perform NLP tasks equally well in all high-resource languages, regardless of the exact resource level.

Comparing the ECE and MCE results for each language in both tasks also shows no great disparity among the languages. No language has the best or worst ECE or MCE in both tasks, indicating that, among high-resource languages, a larger number of resources does not help ChatGPT to better predict the accuracy of its answers.

Comparing the results in the two tasks shows that ChatGPT has a harder time answering CSR questions than performing sentiment analysis. This is suggested by the higher accuracy and lower ECE in the latter. However, compared to other large language models such as XLM-R, ChatGPT seems to be more accurate in CSR and less in SA. This suggests that ChatGPT's worse performance in CSR is due to the task being overall harder, and not to incapacity. Nonetheless, this does not explain why the average confidence in CSR is higher. On the other hand, it appears that ChatGPT is not as skilled in sentiment analysis compared to other state-of-the-art models. Moreover, analysing the standard deviation suggests that regardless of the task, ChatGPT's answers and its confidence in them vary only slightly.

Lastly, English having the best accuracy in both tasks as well as the best ECE for the CSR task suggests that a significant difference in performance among high-resource languages might appear with more or harder tasks.

\subsection{Confidence Calibration}

The additional request to give a confidence value for every answer also led to interesting findings about ChatGPT. Firstly, the range of confidences: ChatGPT never gave a confidence value below 50\%. This is a great issue as it points towards ChatGPT not being aware of its limits. When ChatGPT is faced with a complex task, it should be able to tell its users that its answers may be inaccurate. Instead, it seems that ChatGPT tends to give a confidence value that makes its answer seem trustworthy although it may not be. 

Secondly, its calibration: across all languages and tasks, the average ECE is 15. This value can be considered acceptable, as it suggests ChatGPT has a general idea of its answers' accuracy. However, when considered with the calibration plots, it becomes more worrisome. Although the ECE uses an absolute difference, Figures \ref{sa_cal_plot} and \ref{csr_cal_plot} show that the accuracy is mostly lower than the confidence, suggesting that ChatGPT is often overconfident in its abilities.

Overall, considering the range of confidences and their calibration, it seems that ChatGPT’s confidence in its answers cannot be trusted and should at best be considered as an approximate higher-bound estimate.

\subsection{Limitations}

In the process of studying ChatGPT’s performance and awareness in different high-resource languages, some limitations were encountered. These limitations, which mainly concern the data,  need to be acknowledged to gain a more comprehensive perspective of the results and to highlight areas of improvement for future research.

Firstly, the amount of data. As the data needed to be gathered manually, the amount of data used was quite limited. This caused multiple bins to have very small sizes and this may have influenced the results. With the OpenAI API now available, gathering larger amounts of data should be easier. 

Secondly, the number of tasks. Although the tasks selected show similar results regarding the languages’ performances, using two tasks is not sufficient to generalise about ChatGPT’s abilities confidently. Repeating the experiment with more NLP tasks would potentially strengthen the conclusions. 

Lastly, the translated data of the CSR tasks. As explained by the creators of the dataset \citep{lin2021common}, the data in non-English languages was translated using machine translation and its quality was ensured through an automatic evaluation. However, upon close inspection from native speakers, it appears that some questions or answers contain mistakes that hinder their clarity. This might have influenced ChatGPT to perform worse in non-English languages in the CSR task. 

\section{Conclusions}\label{sec:conclusions}
In this paper, two research questions regarding ChatGPT were defined as the goal of the research. Firstly, whether ChatGPT is equally accurate in high-resource languages, and secondly whether ChatGPT's confidence in its answers is well-calibrated. In order to answer these two questions, ChatGPT was asked to perform 2 NLP tasks (sentiment analysis and common sense reasoning) in 5 high-resource languages (English, French, Italian, Spanish and German) and to give a confidence level for each of its answers.

The results showed that all the languages achieve similar accuracy in both tasks, with no significant difference between them, therefore providing a positive answer to the first research question. The range of confidence levels given, together with the calibration plot and the expected calibration error, suggests that ChatGPT does not know when it lacks knowledge and that it is often overconfident. This answers negatively the second research question.
ChatGPT is equally accurate in high-resource languages but its confidence in its answers is not well calibrated.

With regard to the performance in the two NLP tasks, the comparison to the XLM-R model suggests that ChatGPT is more suited for common sense reasoning than for sentiment analysis.

\subsection{Future Research}
This study provides valuable insights into ChatGPT’s multilingual abilities and confidence calibration. However, as new GPT models get released and gain more and more users, it is fundamental to keep studying their strengths and limitations. The findings of this paper can be used as a basis for a multitude of research areas. 

The same experiment could be repeated on newer models, such as GPT4, to assess whether the performance between languages changes, whether they give a wider range of confidence levels or whether they have better confidence calibration. 

Finally, more NLP tasks or high- to medium-resource languages could be added. Adding tasks of different degrees of difficulty would give the findings a broader generalisation and would confirm whether all high-resource languages perform equally or whether the performance depends on the task. Adding more languages would help assess the resource level at which languages start performing significantly worse than English.

\bibliographystyle{ACM-Reference-Format}
\bibliography{acmart}

\clearpage
\appendix
\section{Decision Between Using English or Translated Prompts} \label{mcnemar}
In order to decide whether to use English prompts or translated prompts a small experiment was conducted. For each language, the first 30 data points of the sentiment analysis task were selected. ChatGPT was instructed to perform sentiment analysis both in English and in the language of the datapoints. The accuracy when using the English prompt or the translated prompts was then computed and can be found in Table \ref{tab:trans}. 

\begin{table}[h]
\centering
    \caption{Accuracy in the first 30 prompts of the sentiment analysis task when using English or translated prompts. German shows a significant difference in accuracy between the two options.}
    \label{tab:trans}
    \begin{tabular}{lll}
       \toprule
       Language & English  & Translated  \\
       \midrule
       English & 76.7 &  N/A \\ 
       French & 73.3 & 73.3  \\ 
       Italian & 76.7 & 76.7  \\ 
       Spanish & 56.7 & 66.7 \\ 
       German & 56.7 &  70.0 \\ 
       \bottomrule
    \end{tabular}
\end{table}

Since German showed the greatest difference in accuracy, it was selected for a McNemar test of significance. The McNemar test determined that there was a statistically significant difference in the accuracy of ChatGPT between using an English or German prompt, $p=0.03$. Since at least one language performed significantly better when using a translated prompt, it was decided to use translated prompts for all languages and tasks.

\section{Translated Prompts} \label{translated}
\subsection{Sentiment Analysis}
\begin{itemize}
    \item \textbf{French: }Bonjour, j'aimerais que tu fasses une analyse de sentiment sur 15 phrases françaises. Classe chaque phrase comme "Positive" si elle exprime un sentiment positif, "Negative" si elle exprime un sentiment négatif ou "Neutral" si elle est neutre." Aussi, évalue ta confiance dans  la réponse que tu as donnée entre 0 et 100 \%. La réponse doit être une liste au format "Sentiment (confiance\%)". Réponds "compris" si tu  as compris. La liste des phrases suivra.
    \item \textbf{Italian: }Ciao, vorrei che tu eseguissi un'analisi del sentimento su 15 frasi italiane. Classifica ogni frase come “Positive” se esprime un sentimento positivo, “Negative” se esprime un sentimento negativo o “Neutral” se è neutra. Inoltre, valuta quanta fiducia hai nella risposta che hai dato tra 0 e 100\%. La risposta dovrebbe essere un elenco con il formato "Sentimento (fiducia\%)". Rispondi "capito" se hai capito. Seguirà l'elenco delle frasi.
    \item \textbf{Spanish: }Hola, me gustaría que realizaras un análisis de sentimiento de 15 frases en español. Clasifica cada frase como “Positive” si expresa un sentimiento positivo, “Negative” si expresa un sentimiento negativo o “Neutral” si es neutra. Además, puntúa tu confianza en la respuesta que has dado entre 0 y 100\%. La respuesta debe ser una lista con el formato "Sentimiento (confianza\%)". Contesta “entendido” si lo has entendido. A continuación aparecerá la lista de frases.
    \item \textbf{German: }Hallo, ich möchte Dich bitten, eine Stimmungsanalyse für 15 deutsche Sätze durchzuführen. Ordne jeden Satz als "Positive" ein, wenn er eine positive Stimmung hat, "Negative", wenn er eine negative Stimmung hat oder "Neutral", wenn er neutral ist. Bewerte bitte auch Dein Vertrauen in  Deine Antwort zwischen 0 und 100\%. Die Antwort muss eine Liste im Format "Stimmungsart (Vertrauen\%)" sein. Antworte mit "verstanden", wenn Du verstanden hast. Es folgt eine Liste von Sätzen.
\end{itemize}

\subsection{Common Sense Reasoning}
\begin{itemize}
    \item \textbf{French: } Bonjour, j'aimerais que tu répondes à 15 questions de bon sens à choix multiple en français. Pour chaque question, donne la lettre qui correspond à la bonne réponse. Aussi, évalue ta confiance dans la réponse que tu as donnée entre 0 et 100 \%. La réponse doit être une liste au format "Lettre (confiance\%)". Réponds "compris" si tu as compris. La liste des questions suivra.
    \item \textbf{Italian: }Ciao, vorrei che tu rispondessi a 15 domande di senso comune a scelta multipla in italiano. Per ogni domanda, scegli la lettera che corrisponde alla risposta corretta. Inoltre, valuta quanta fiducia hai nella risposta che hai dato tra 0 e 100\%. La risposta dovrebbe essere un elenco con il formato "Lettera (fiducia\%)". Rispondi "capito" se hai capito. Seguirà l'elenco delle domande.
    \item \textbf{Spanish: }Hola, me gustaría que respondieras a 15 preguntas de sentido común de opción múltiple en español. Para cada pregunta, por favor, indique la letra que corresponde a la respuesta correcta. Además, puntúa tu confianza en la respuesta que has dado entre 0 y 100\%. La respuesta debe ser una lista con el formato "Letra (confianza\%)". Contesta “entendido” si lo has entendido. A continuación aparecerá la lista de preguntas.
    \item \textbf{German: }Hallo, ich möchte Dich bitten, 15 Multiple-Choice-Fragen des gesunden Menschenverstands auf Deutsch zu beantworten. Bitte gib für jede Frage den Buchstaben an, der der richtigen Antwort entspricht. Bewerte bitte auch Dein Vertrauen in Deine Antwort zwischen 0 und 100\%. Die Antwort muss eine Liste im Format "Buchstabe (Vertrauen\%)" sein. Antworte mit "verstanden", wenn Du verstanden hast. Es folgt die Liste von Fragen.
\end{itemize}

\FloatBarrier

\section{Sample ChatGPT Queries and Responses} \label{examp}
In this section, for each language, six examples of ChatGPT performing sentiment analysis and common sense reasoning are presented. For each language and task, three correct and three incorrect answers were randomly sampled. The tables present the tweet or question (and options) ChatGPT was given, ChatGPT's answer with its confidence level as well as the correct answer. In every table, the first three rows present correct answers and the last three present wrong answers.

\subsection{Sentiment Analysis}

\begin{table*}[]
\centering
    \caption{Three correct and three incorrect answers ChatGPT gave while performing Sentiment Analysis in English.}
    \label{tab:sa_en}
    \begin{tabular}{p{0.65\textwidth} p{0.15\textwidth} p{0.1\textwidth}}
       \toprule
       Tweet & Answer   & Correct  \\
       \midrule
       @user on Comedy Central tonight @ 9...I'm a Grown Little Man and Seriously Funny...I'll be glued to the tv & Positive (80\%) &  Positive \\ 
       "Yo, if you paid \$57-\$77 to go see Eric Church at the State Fair...may the Lord have mercy on your souls. " & Negative (80\%) &  Negative \\ 
      "So who's next? Maduro? A little young, perhaps, but it is Venezuela.... " & Neutral (75\%) &  Neutral \\ 
      \midrule
       looks like I'm done with Kpop guys & Negative (89\%) &  Neutral \\ 
       If I'm Dustin Johnson I wouldn't care about my Sunday meltdowns going home to Paulina. \#sheesh  & Positive (80\%) &  Negative\\ 
       I always thought Prince was the 18th best songwriter of all time but it's nice of Rolling Stone to confirm that. &  Neutral (75\%) & Positive \\ 
        \bottomrule
    \end{tabular}
\end{table*}

\begin{table*}[]
\centering
    \caption{Three correct and three incorrect answers ChatGPT gave while performing Sentiment Analysis in French.}
    \label{tab:sa_en}
    \begin{tabular}{p{0.65\textwidth} p{0.15\textwidth} p{0.1\textwidth}}
        \toprule
       Tweet & Answer   & Correct  \\
       \midrule
       "Sévère réquisitoire de l'ex-ministre de l’Écologie, Delphine \#Batho, contre l'action du chef de l’État dans un... http" & Negative (90\%) &  Negative \\
Formation aux rencontres écologiques d'été avec un futur eurodéputé espagnol issu des Indignés. Passionnant! \#ree14 http & Positive (75\%) &  Positive \\
      Fédération des Parcs Naturels Régionaux | @user http & Neutral (65\%) &  Neutral \\
      \midrule
       Aides à l’agriculture biologique : les écologistes mobilisés http & Neutral (70\%) &  Positive \\ 
       Région PACA : le nouvel écosystème de l'export en ordre de marche http  & Positive (75\%) &  Neutral\\ 
       Si on ne peut même plus rêver... \#animaux \#amitié http &  Neutral (65\%) & Negative \\ 
        \bottomrule
    \end{tabular}
\end{table*}

\begin{table*}[]
\centering
    \caption{Three correct and three incorrect answers ChatGPT gave while performing Sentiment Analysis in Italian.}
    \label{tab:sa_en}
    \begin{tabular}{p{0.65\textwidth} p{0.15\textwidth} p{0.1\textwidth}}
        \toprule
       Tweet & Answer   & Correct  \\ 
       \midrule
       "Buongiorno e ""buona scuola"" a tutti... finalmente è proprio il caso di dirlo!" & Positive (80\%) &  Positive \\ 
Dimettiti incapace! @user http & Negative (70\%) &  Negative \\ 
      Lettera al premier mario monti Corriere della Sera Inviato da iPad http & Neutral (60\%) &  Neutral \\ 
      \midrule
       "Metti un ""like"" per la \#scuola paritaria \#iniziativa http \#labuonascuola" & Positive (80\%) &  Neutral \\ 
       "@user si certo, ma questa definizione passa spesso in secondo piano a seconda di chi viene indagato"  & Neutral (70\%) &  Negative\\ 
       Batteria 2\%...buonanotte a tutti &  Negative (90\%) & Neutral \\ 
        \bottomrule
    \end{tabular}
\end{table*}

\begin{table*}[]
\centering
    \caption{Three correct and three incorrect answers ChatGPT gave while performing Sentiment Analysis in Spanish.}
    \label{tab:sa_en}
    \begin{tabular}{p{0.65\textwidth} p{0.15\textwidth} p{0.1\textwidth}}
        \toprule
       Tweet & Answer   & Correct  \\ 
       \midrule
       "@user hombreee, poco a poco, los 5 primeros minutos que montes en bici te sabrán a gloria  así que ánimo que todo llega"" & Positive (90\%) &  Positive \\ 
@user pero eso no es grasioso & Negative (80\%) &  Negative \\ 
      "Aveces cuando leo sus tuits y no entiendo algo, me dan ganas de preguntarles para entender mejor. Pero soy tímido y me quedo con la duda" & Neutral (75\%) &  Neutral \\
      \midrule
       Debo dejar de hacer bromas porque cuando hable en serio no me creerán & Negative (70\%) &  Neutral \\ 
       Bienvenida a los juegos mentales  & Negative (80\%) &  Positive\\ 
       Esta para unos buenos mates y galletitas &  Positive (90\%) & Neutral \\ 
        \bottomrule
    \end{tabular}
\end{table*}

\begin{table*}[]
\centering
    \caption{Three correct and three incir answers ChatGPT gave while performing Sentiment Analysis in German.}
    \label{tab:sa_en}
    \begin{tabular}{p{0.65\textwidth} p{0.15\textwidth} p{0.1\textwidth}}
        \toprule
       Tweet & Answer   & Correct  \\ 
       \midrule
       "@user Weitermachen, hopp hopp. Steht noch genug aus ;)"& Positive (80\%) &  Positive \\ 
"RT @user: Entrüstung über Merkel-Abhöraktion. Berechtigt? Ja, natürlich! Ja!! Aber Front im Cyberwar verläuft längst woanders http" & Negative (75\%) &  Negative \\ 
      RT @user: Sei nicht zu müde! Nächster Halt: Romantik mit \#Flirtbotschaft http \#München & Neutral (70\%) &  Neutral \\ 
      \midrule
       Ich habe ein @user-Video positiv bewertet: http Let's (schlechte) Parodie: 24SpeckTV & Neutral (50\%) &  Positive \\ 
       @user große katastrophe :P  & Negative (65\%) &  Positive\\ \hline
       "Gute Besserung mutee {} RT ""@user: Endlich muss ich zum Arzt gehen :|""" &  Positive (75\%) & Negative \\ 
        \bottomrule
    \end{tabular}
\end{table*}

\FloatBarrier
\subsection{Common Sense Reasoning}

\begin{table*}[]
\centering
    \caption{Three correct and three incorrect answers ChatGPT gave while performing Common Sense Reasoning in English.}
    \label{tab:sa_en}
    \begin{tabular}{p{0.7\textwidth} p{0.1\textwidth} p{0.1\textwidth}} 
        \toprule
       Question & Answer   & Correct  \\
       \midrule
       What castle is built upon Castle Rock in Scotland? \\A: capturing pawn. B: london. C: edinburgh. D: germany. E: europe. & C (90\%)  &  C  \\ 
       John didn't like eating the hamburger.  It made him feel what?\\ A: enjoy. B: naus
       ea. C: sad. D: death. E: satisfaction. & B (95\%) &  B  \\ 
      Everyone is special. Everyone has what?\\ A: feelings. B: values. C: unique personality. D: experiences. E: different standards. & C (95\%) &  C \\ 
      \midrule
       Some people really loved beer, they call it what gold? \\A: alcohol in. B: liquid. C: intoxicating. D: hair of the dog. E: harmful. & D (70\%) &  B \\ 
       38. Stopping being married to her allowed him to again pursue his dreams, it brought him what? \\A: depression. B: wrong. C: relief. D: rememberance. E: pleasure.  & C (80\%) &  E\\ 
       After the guy is successful in cashing in his check, what does he feel? \\ A: sad. B: quitting. C: extra money. D: leave. E: great joy. &  C (85\%) & E \\ 
        \bottomrule
    \end{tabular}
\end{table*}

\begin{table*}[]
\centering
    \caption{Three correct and incorrect answers ChatGPT gave while performing Common Sense Reasoning in French.}
    \label{tab:sa_en}
    \begin{tabular}{p{0.7\textwidth} p{0.1\textwidth} p{0.1\textwidth}}
        \toprule
       Question & Answer   & Correct  \\ 
       \midrule
       Où les élèves collent-ils des chewing-gums à l'école?\\ A: livres. B: trottoir. C: fontaine d'eau. D: étagère. E: films. & C (80\%)  &  C  \\ 
    Sauter par-dessus une corde est une forme de quoi ? \\A: maux de tête. B: fun. C: se réchauffer. D: exercice. E: brisé. & D (90\%) &  D  \\ 
    Où les étudiants sont-ils susceptibles d'utiliser une structure de stationnement ? \\A: grande ville. B: centre commercial. C: campus universitaire. D: derrière le garage.. E: chicago. & C (75\%) &  C \\ 
        \midrule
       Que doit faire un enseignant pour ses élèves ? \\ A: une meilleure connaissance. B: groupe d'étudiants. C: préparer le déjeuner. D: test de temps. E: énoncer les faits.. & A (85\%) &  E \\ 
       Quel est l'endroit idéal pour avoir un chat ?\\ A: chaise confortable. B: walmart. C: société humaine. D: étage. E: appui de fenêtre.  & E (80\%) &  C\\ 
       Où peut-on voir un point de repère entouré de plusieurs kilomètres de plantes, si ce n'est dans un parc ?\\ A: campagne. B: parc national. C: désert. D: carte. E: ville. &  B (80\%) & A \\ 
       \bottomrule
    \end{tabular}
\end{table*}

\begin{table*}[]
\centering
    \caption{Three correct and three incorrect answers ChatGPT gave while performing Common Sense Reasoning in Italian.}
    \label{tab:sa_en}
    \begin{tabular}{p{0.7\textwidth} p{0.1\textwidth} p{0.1\textwidth}}
        \toprule
       Question & Answer   & Correct  \\ 
       \midrule
       Qual è il solito motivo per cui la gente si diverte a giocare?\\ A: felicità. B: intrattenimento. C: competitività. D: rabbia. E: vincente. & B (90\%)  &  B  \\ 
    Gli esseri umani imparano l'attualità da quale articolo di stampa? \\A: domandare l'autorità. B: leggere i giornali. C: sciogliere. D: agire. E: dizionario. & B (80\%) &  B  \\ 
    Dove si sentirebbe più libero un furetto?\\ A: all'aperto. B: carolina del nord. C: nord america. D: gran bretagna. E: aula. & A (90\%) &  A \\
    \midrule
       Dove sarebbe improbabile vedere delle impronte a causa dell'assenza di persone?\\ A: terra. B: tappeto. C: camino. D: tetto. E: superficie di terra. & A (85\%) &  D \\ 
       Aveva una relazione, cosa è finito con la scoperta del marito? \\A: relazione. B: matrimonio. C: fedeltà. D: amore. E: divorzio.  & E (75\%) &  B\\ 
       Sarah ama dare assistenza.  Non chiede nulla in cambio.  A cosa probabilmente Sarah tiene di più?\\ A: i suoi capelli. B: circostanze migliori. C: felicità. D: imparare nuove competenze. E: persone aiutate. &  C (85\%) & E \\ 
       \bottomrule
    \end{tabular}
\end{table*}

\begin{table*}[]
    \small
    \centering
    \caption{Three correct and three incorrect answers ChatGPT gave while performing Common Sense Reasoning in Spanish.}
    \label{tab:sa_en}
    \begin{tabular}{p{0.7\textwidth} p{0.1\textwidth} p{0.1\textwidth}} 
        \toprule
       Question & Answer   & Correct  \\ 
       \midrule
       ¿Dónde podrías averiguar sobre una vaca que no existe? \\A: al aire libre. B: libro de cuentos. C: campo. D: Tienda de juguetes. E: corral. & B (90\%)  &  B  \\ 
    El empleado preguntó qué tipo de carne y pan quería el cliente, y eligieron tener un qué? \\A: bifurcación. B: congelador. C: horno. D: sartén. E: sándwich de jamón. & E (80\%) &  E  \\ 
    ¿Qué es lo que Dios nunca hará según la religión? \\ A: cualquier cosa. B: hacer milagros. C: juzgar a la gente. D: Juez de hombres. E: en todas partes. & C (85\%) &  C \\ 
    \midrule
       ¿Qué hará un agente inmobiliario sin escrúpulos para que un listado parezca más atractivo? \\A: contratar una banda. B: exagerar. C: encantador. D: engañar. E: mentira. & B (70\%) &  E \\ 
       Estaba encontrando información a través de la meditación y el yoga, ¿qué estaba buscando? \\ A: obtener respuestas. B: úlceras. C: poder. D: felicidad. E: respeto.  & A (90\%) &  D\\ 
       Kuwait y otros países de la región no siempre se asocian inmediatamente con ella, pero están situados en donde?\\ A: estados del golfo. B: oriente medio. C: arabia. D: kuwait. E: asia &  B (85\%) & E \\ 
       \bottomrule
    \end{tabular}
\end{table*}

\begin{table*}[]
    \small
    \centering
    \caption{Three correct and three incorrect answers ChatGPT gave while performing Common Sense Reasoning in German.}
    \label{tab:sa_en}
    \begin{tabular}{p{0.7\textwidth} p{0.1\textwidth} p{0.1\textwidth}}
        \toprule
       Question & Answer   & Correct  \\ 
       \midrule
       Ein Arzt kann ein Antibiotikum in vielen Formen verschreiben, welcher Typ könnte eine Gel-Beschichtung haben? \\A: Kapsel. B: Medizinschrank. C: bakterielle Infektion heilen. D: vorgeschrieben. E: Apotheke. & A (80\%)  &  A  \\ 
    Wo kann man einen Tennisplatz für Nicht-Profis finden?\\ A: Erholungszentrum. B: Country-Club. C: Sportverein. D: College-Campus. E: zoo. & A (90\%) &  A  \\ 
    Welcher Teil eines Grundstücks darf neben einer Einfahrt liegen?\\ A: Auto weiterfahren. B: Nachbarschaft. C: Hof. D: Unterteilung. E: Vorort. & C (85\%) &  C \\ 
    \midrule
       Warum sollte eine Person an ihrem Arbeitsplatz bleiben, ohne bezahlt zu werden? \\A: Gott sei Dank. B: Schlange stehen. C: gerne arbeiten. D: Hilfe anbieten. E: Freiwillige. & E (70\%) &  C \\ 
       Wo werden Sie wahrscheinlich Sterne aus dem Fenster sehen? \\A: Spaceshuttle. B: Gebäude. C: auf dem Rasen. D: Haus eines Freundes. E: Wand.  & D (75\%) &  A\\ 
       Warum verdienen Sie kein Geld, während Sie zu Mittag essen?\\ A: die Arbeit einstellen. B: Nahrung zu sich nehmen. C: Essen besorgen. D: Nahrung finden. E: zu voll.  &  B (90\%) & A \\ 
       \bottomrule
    \end{tabular}
\end{table*}

\FloatBarrier

\section{Zero Shot Chain-of-Thought Reasoning Results}

\subsection{CoT Prompts} \label{CoTPrompt}
\subsubsection{Sentiment Analysis}
\begin{itemize}
    \item \textbf{Baseline: }Hello, I would like you to perform sentiment analysis on an English sentence. Please classify it as “Positive”, “Negative”, or “Neutral” according to the sentiment it expresses. Moreover, please rate your confidence in the answer you gave between 0 and 100\%. The answer should have the format “Sentiment (confidence\%)”. The sentence will follow.
    \item \textbf{Zero Shot Chain of Thought: }Hello, I would like you to perform sentiment analysis on an English sentence. Please classify it as “Positive”, “Negative”, or “Neutral” according to the sentiment it expresses. Moreover, please rate your confidence in the answer you gave between 0 and 100\%. The answer should have the format “Sentiment (confidence\%)”. Please explain your reasoning step by step. The sentence will follow.
\end{itemize}
\subsubsection{Common Sense Reasoning}
\begin{itemize}
    \item \textbf{Baseline: }Hello, I would like you to answer a common sense multiple-choice question in English. Please give the letter that corresponds to the correct answer. Moreover, please give your confidence in the answer you gave between 0 and 100\%. The answer should have the format “Letter (confidence\%)”. The question will follow.
    \item \textbf{Zero Shot Chain of Thought: }Hello, I would like you to answer a common sense multiple-choice question in English. Please give the letter that corresponds to the correct answer. Moreover, please give your confidence in the answer you gave between 0 and 100\%. The answer should have the format “Letter (confidence\%)”. Please explain your reasoning step by step. The question will follow.    
\end{itemize}

\FloatBarrier
\subsection{ChatGPT Queries and Responses}
\label{CoTResults}

\begin{table}[]
    \centering
    \caption{ChatGPT's Answer and Confidence comparison for Standard Prompt and Zero Shot Chain of Thought prompt in Sentiment Analysis.}
    \label{tab:CoTSA}
    \begin{tabular}{p{0.50\textwidth} p{0.15\textwidth} p{0.15\textwidth} p{0.1\textwidth}}
        \toprule
        Tweet & Baseline& ZS CoT & Correct \\
        \midrule
        Pope concelebrates Mass with Armenian Patriarch: History was made on Monday when Pope Francis concelebrated mo.. & Neutral 85\% & Neutral 85\% & Positive\\ 
        In one weeks time I managed to binge watch the entire 4th season of Gilmore Girls and write my Stress management plan/paper \#Multitasker
        & Positive 80\% & Positive 80\% & Neutral\\
        "Those criticizing her for the ""I thought I was the only one allowed to **** Verlander"" need to find a hobby, and maybe a job. ""
        & Negative 75\% & Negative 70\% & Neutral\\
        @user okay I'll admit PC may be better but PS4 is just my preference and always will be. I just like console lol. 
        & Neutral 75\% & Positive 65\% & Positive\\
        The only reason Dustin Johnson didn't play  well is because it's a Sunday.  Luckily for him the final round is on Monday.  \#TheOpen2015
        & Neutral 70\% & Neutral 60\% & Positive\\
        \bottomrule
    \end{tabular}
\end{table}

\begin{table}[]
    \centering
    \caption{ChatGPT's Answer and Confidence comparison for Standard Prompt and Zero Shot Chain of Thought prompt in Common Sense Reasoning.}
    \label{tab:CoTSA}
    \begin{tabular}{p{0.50\textwidth} p{0.15\textwidth} p{0.15\textwidth} p{0.1\textwidth}}
        \toprule
        Question & Baseline& ZS CoT & Correct \\
        \midrule
        Some people really loved beer, they call it what gold? \\A: alcohol in. B: liquid. C: intoxicating. D: hair of the dog. E: harmful. & D 80\% & C 95\% & B\\
        Where is a fox likely to be caught?\\ A: tennessee. B: inside joke. C: the forrest. D: grassy field. E: england.
        & E 80\% & E 90\% & D\\
        Some celebrities find going public useful after being caught at what? \\A: high heel shoes. B: getting high. C: problems. D: press coverage. E: wide acceptance.
        & D 85\% & D 95\% & B\\
        Stopping being married to her allowed him to again pursue his dreams, it brought him what?\\ A: depression. B: wrong. C: relief. D: rememberance. E: pleasure.
        & C 85\% & C 90\% & E\\
        If you don't use a towel after getting wet, your body may do what? \\A: not dry. B: shrinkage. C: become cold. D: shiver. E: get melted.
        & C 80\% &C 95\% & D\\
        \bottomrule
    \end{tabular}
\end{table}

.
\end{document}